# A Comparative Analysis of Multiple Methods for Predicting a Specific Type of Crime in the City of Chicago

Deborah Djon, Jitesh Jhawar, Kieron Drumm, and Vincent Tran

*School of Computing, Dublin City University, Dublin, Ireland*

Researchers regard crime as a social phenomenon that is influenced by several physical, social, and economic factors. Different types of crimes are said to have different motivations. Theft, for instance, is a crime that is based on opportunity, whereas murder is driven by emotion. In accordance with this, we examine how well a model can perform with only spatiotemporal information at hand when it comes to predicting a single crime. More specifically, we aim at predicting theft, as this is a crime that should be predictable using spatiotemporal information. We aim to answer the question: "How well can we predict theft using spatial and temporal features?". To answer this question, we examine the effectiveness of support vector machines, linear regression, XGBoost, Random Forest, and k-nearest neighbours, using different imbalanced techniques and hyperparameters. XGBoost showed the best results with an F1-score of 0.86.

CCS Concepts: • **Computing methodologies** → **Machine Learning** → **Learning paradigms** → **Supervised learning** → **Supervised learning by classification**

Additional Key Words and Phrases: crime prediction, logistic regression, support vector machine, XGBoost, k-nearest neighbours, Random Forest

## 1. RELATED WORK

### 1.1 Motivations

Historically, attempts have been made to use technology to predict crime trends, primarily with the use of traditional regression models. However, though such models have been found to be useful at highlighting the contribution that features of a crime have to a prediction, they are not considered as being optimal for prediction [4]. Due to this limitation, further attempts have been made to use advanced machine learning models to make predictions. These models have been shown to be more accurate at prediction but limited in their transparency and interpretability [4]. A number of research studies have been carried out in recent years in an attempt to rectify this particular shortcoming: in one instance, we have the work of [4], whose criticisms of traditional crime prediction models have led them to suggest the use of the XGBoost algorithm to train such a model. The argument here was that this model, paired with a Shapley Additive Explanation (SHAP) tree, would provide a considerably higher level of transparency and interpretability [4].

There is also the belief that crime is a social phenomenon — one that is influenced by a number of physical, social, and economic elements, and that the relationship between such elements and the types of crime to which they contribute is one that is "beyond human perception" [2]. This belief has influenced the work of [2], who believe that we must look past the usual spatiotemporal features that are commonly considered when training crime prediction models. Instead, they believe that specific crimes must be categorised and the relationships between these crimes and certain environmental factors studied in order to gain a deeper understanding of the correlation between these factors and the crimes that they influence [2]. To do so, they suggested the use of an artificial neural network (ANN) [2].

### 1.2 Target Crime(s)

In previous studies, some researchers have chosen to focus on a specific type of crime, whereas others have attempted to predict a number of different types of crime in a specific city or region. In the case of [2], a focus was put on theft, as they believed it to be most affected by spatial properties. This was due to the belief that those who commit theft will normally target anything available to them, as opposed to murder, which is normally driven by emotion and targeted at a specific person or people [2]. To train a model to predict the occurrence of theft, the necessary data was pulled from a dataset containing recordings of past crimes that had occurred in the Dongjak district of Seoul between 2004 and 2015 [2].

Similarly, in the work of [4], a focus was also put on theft, but in this case, specifically on public theft, due to an observation made on the dataset used that showed that public theft constituted 40% of all crime incidents in the target area from 2017 to 2020 — the target area, in this case, being the XT paichusuo (police station or beat) of ZG (Zigong) city in Sichuan province,

China [4]. As referenced above, the dataset used was a public record of all public thefts that had occurred in the aforementioned paichusuo from 2017 to 2020 [4].

Although [2] and [4] made a point of focusing on some form of theft in their studies, [5] chose instead to focus on analysing and predicting the top 10 most frequently occurring types of crime in the city of San Francisco — which accounted for 97% of the overall crime in the area [5]. For this study, a Kaggle dataset containing records of crimes from the city of San Francisco was used, which ranged from 2003 to 2015 [5].

1.3 Feature Selection

One of the biggest obstacles to overcome when training any machine learning model is selecting the features to be used, whether it be by intuitive or statistical means. In the field of crime prediction, many approaches have been taken to find the optimal method for finding these relevant features. In the case of [2], a great deal of importance was placed on spatiotemporal features due to the inspiration that they took from Oscar Newman's "Defensible Space", which presented the possibility that crime could be prevented through urban design, and as such, could be influenced by environmental features [6]. With this in mind, the most important environmental features were isolated with the use of multiple regression [2].

Similar to [2], the features chosen by [4] were inspired by both crime pattern theory and routine activity theory. Crime pattern theory focuses on places and opportunities, suggesting that offenders will always tend to choose locations with which they are familiar when searching for potential targets [2, 7], whereas routine activity theory argues that the "convergence of motivated offenders, suitable targets/victims, and the absence of capable guardians at time and space is needed for a crime event to happen" [4, 6]. All features chosen by [4] were selected with these beliefs in mind, and their overall contributions were subsequently evaluated in all trained models using Shapley additive explanation (SHAP).

Unlike the work of [2] and [4], [5] chose to follow a more intuitive approach to feature selection, utilising exploratory data analysis (EDA) in order to isolate the features that they ultimately deemed to be irrelevant "due to their nature" [5].

1.4 Imbalanced Data

When dealing with large datasets, one is bound to come across imbalances. Imbalances can cause issues when attempting to train a classification model, as the majority of machine learning algorithms tend to assume, by default, that the data with which they are dealing is balanced [13]. This assumption results in the outputs of the trained models being biased and skewed towards the majority class [13].

Some researchers have suggested that the most effective method of dealing with these imbalances is via oversampling, using the Synthetic Minority Oversampling Technique (SMOTE) or the arguably improved form of SMOTE: Adaptive Synthetic (ADASYN) [13]. Past research has shown that, when comparing the effectiveness of four separate models, trained using logistic regression, support vector machine, Random Forest, and XGBoost, the application of either SMOTE or ADASYN resulted in a significant increase in the F1-score and AUC-ROC [13]. For example, in the case of Random Forest, a model trained on an untouched dataset was shown to produce an F1-score of 0.18 and an AUC-ROC of 0.55, whereas a model trained on a newly balanced dataset was shown to produce an F1-score of 0.91 and an AUC-ROC of 0.91 [13]. Though these researchers acknowledged the shortcomings of SMOTE, such as its tendency to oversample minority instances with a uniform likelihood or the fact that it tends to intensify the noise [13], they still concluded that the use of one of these techniques resulted in a better model overall [13].

Others have proposed the use of ensemble algorithms in order to resolve this issue. For example, [14] proposed the use of an ensemble algorithm that utilised undersampling, cost-sensitive learning, bagging, and support vector machine (SVM), in order to train an effective model on imbalanced medical data. By doing so, they were able to outperform common ensemble algorithms, such as ADABoost, EasyEnsemble, and Random Forest, as well as commonly used algorithms, such as k-nearest neighbours (KNN) and logistic regression [14], whereas [15] stressed the importance of placing value on samples that exist near the decision boundaries between classes, as such samples may contain discriminative information. With this in mind, they proposed a combination of "Borderline-SMOTE" and an ensemble of SVMs, aptly named: "Bagging of

Extrapolation Borderline-SMOTE SVM (BEBS)" [15].

1.5 Implementation

Though machine learning can be very useful in making general predictions, there has been some debate over which method is optimal for crime prediction. For example, in the case of [2], the researchers here chose to cluster occurrences of theft using the k-modes algorithm, an unsupervised machine learning algorithm that is especially useful when working with categorical data [2], before then using an artificial neural network (ANN) to learn how to predict the occurrences of said clusters. In the work of [4], the accuracies of logistic regression, decision trees, and the Random Forest algorithm were compared before they ultimately chose to use the XGBoost algorithm, as it was shown to have the best performance [4], and in the work of [5], the accuracies of Naive Bayes, Random Forest, and gradient-boosted decision trees were compared [5].

In terms of results, [2], while recognising that they had only taken environmental factors into account, found that there were typically four types of theft that occurred in the target area: theft in commercial facilities, home intrusions, vehicle-related theft, and "other" [2]. For each of the four clusters of features that were previously identified, the researchers were able to train an ANN to predict theft in the target area with an $R^2$ of 0.262, 0.210, 0.245, and 0.148, respectively. Ultimately, they concluded that the correlation between thefts and local environmental factors showed a necessity for an environmental criminology approach to theft prevention [2]. [4], using the XGboost algorithm, was able to achieve an accuracy of 0.91 with its training set, 0.89 with its verification set, and an AUC-ROC of 0.586 [4]. Though [4] recognised that their study had some shortcomings in that it had only been tested in one city, they still found that, interestingly, more theft occurred in areas with a higher ambient population, aged between 25 and 44, and that the size of this population had an effect on the output of any trained models [4]. Lastly, [5] was able to achieve an accuracy of 65.82% (Naive Bayes), 63.43% (Random Forest), and 98.5% (gradient-boosted decision trees) with the models that they trained, ultimately concluding that gradient-boosted decision trees were the ideal choice for crime prediction [5].

## 2. METHODOLOGY

For each section of our analysis, a methodology was followed when working on the data — in our case, the data was preprocessed by cleaning and handling missing values, outliers, and duplications. Exploratory data analysis (EDA) was then conducted to understand the underlying patterns and relationships in the data. Once this EDA was completed, we proceeded to remove any features that we had identified as having high levels of correlation in order to improve model performance. In some cases, Multiple correspondence analysis (MCA) was performed in order to reduce dimensionality and further identify important features. The resultant dataset (post-reduction) was then scaled in order to normalise all features and to avoid bias towards any particular feature.

Finally, Following similar methods used in most papers we found such as in [3] or [21], a collection of algorithms (linear and logistic regression, support vector machine (SVM), Random Forest, and k-nearest neighbours (KNN)) and XGBoost was applied to our data. Each model was then evaluated using metrics such as accuracy, precision, recall, F1-score, and AUC-ROC (see results below for further information). Once each model had been trained initially, some hyperparameters were tuned with the use of GridSearchCV in order to optimise each model, after which the models were subsequently compared to each other and their overall results recorded.

2.1 Exploratory Data Analysis

An exploratory data analysis of the dataset revealed a significant imbalance in the dataset with regards to the "Primary Type". This is evident in Figure 1, which shows how "Theft", "Battery", and "Criminal Damage" constitute a large portion of the overall crimes that occur in Chicago. The location features that were used contained several outliers that had values of 0. This indicated that the exact location of the crime had not been recorded. During the analysis phase, several features were also found to be correlated. The Illinois Uniform Crime Reporting (IUCR) and Federal Bureau of Investigation (FBI) codes had an R score of 0.86, the "Beat" and "Ward" had an R score of 0.64, and the "Ward" and "District" had an R score of 0.69. The first correlation between the IUCR and FBI codes can be explained by the fact that they both serve the same purpose of being identifiers that are assigned to reports of criminal acts [1]. The latter correlations can be explained in a similar manner, as they are all measures of location. Further inspection of these correlated features, namely the "Beat", "District", and "Year", showed that they had considerably high variance inflation factors (VIFs), above 35 on average.

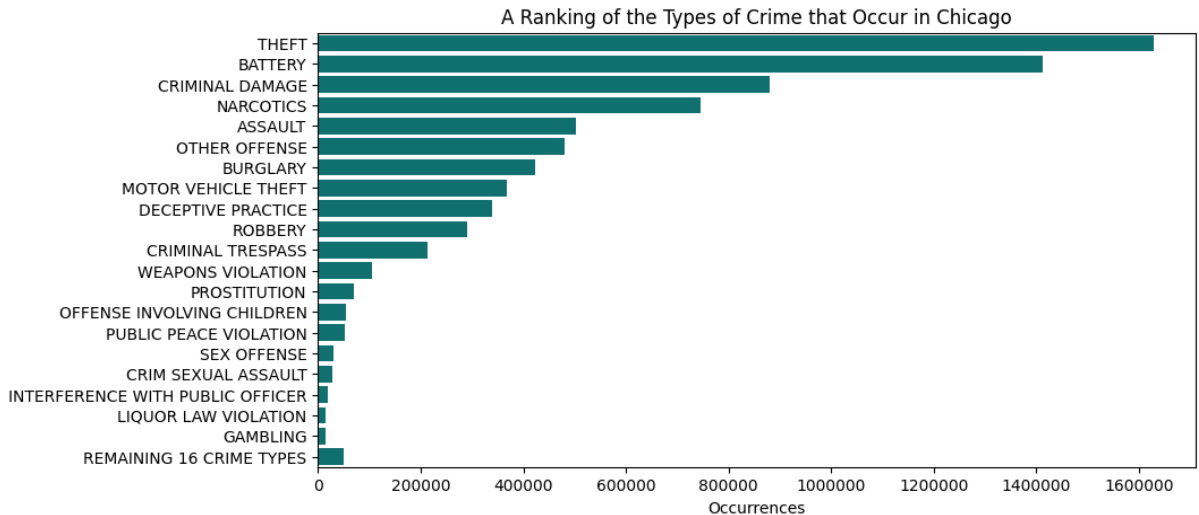

Figure 1. A Ranking of the Types of Crime that Occur in Chicago.

2.2 Data Preparation

As a first step in data preparation, we performed feature engineering by removing less meaningful existing features and creating new features.

We removed features that are recorded after a crime has occurred, as they cannot be used for prediction. These features include identifiers such as 'ID', 'FBI Code', 'IUCR' and 'Case Number', 'Arrest', which logs whether a criminal was arrested or not, and 'Last Updated'. The 'Description' feature was removed as it is redundant. It is a more granular version of our target feature, 'Primary Type'. Similarly, we remove the 'Location' feature as it is analogous to the other spatial descriptor features.

We examined the remaining features for multicollinearity using the Pearson Correlation Coefficient for numerical variables, the Chi-Square test of independence for nominal features, and Spearman's Rank Correlation Coefficient for ordinal variables. Based on these results, we also remove the features 'Latitude' and 'Longitude' as they are highly correlated with the X and Y coordinate features.

Next, we proceeded to remove the 'Date' feature and replace it with a new feature: "Week Number", which represented the number of the Week (out of 52) during which the crime occurred, the month and weekday on which the crime occurred, and the hour at which the crime occurred.

In order to handle any missing values in our dataset, we used a KNN-based approach to impute any missing data, as such an approach has been shown to outperform standard sample and median-based imputation approaches in recent studies [12]. We tried to convert categorical features into one-hot-encoding for models that require linearity. However, our system was not able to computationally live up to the task. For the other models, we used label encoding for ordinal and nominal features. We also shuffled the dataset to prevent potential patterns in the order of the data influencing the model's performance.

In the case of SVM, logistic regression, and KNN, a selection of our features were scaled to fit the normal distribution. Due to processing limitations, some models were trained on a smaller subset of the dataset. We used 125,000 records for the KNN and Random Forest models, 10,000 samples for the SVM model and 100,000 for XG boost.

We used a SMOTE oversampler with 3 K-nearest neighbours and a random undersampler to handle the imbalance in our dataset.

2.3 Models

2.3.1 K-nearest Neighbours

K-nearest neighbours (KNN) is a supervised learning algorithm used for both regression and classification tasks. It makes predictions on given inputs by performing a search for the "k" closest data points, also referred to as neighbours, in the dataset and classifies the latest input based on the class with which it shares the most neighbours. The value of "k" in this algorithm is a hyperparameter that requires tuning based on the problem at hand [9]. KNN is also a non-parametric algorithm, meaning that it does not make any assumptions about the underlying distribution of the data on which it is being trained; it is also relatively easy to understand and implement. However, it can be computationally expensive for large datasets, it can encounter issues with datasets that have high dimensions, and the choice of an optimal value for "k" can have a significant impact on the performance of the algorithm [9]. In this study, a KNN Classifier was used with 14 neighbours and the Manhattan metric to calculate the distance between the possible neighbours. To reduce the dimensionality of our dataset and the computation time, Multiple Correspondence Analysis (MCA) was used, as KNN does not perform well in high dimensions [17].

### 2.3.2 Logistic Regression

Logistic regression is a supervised learning method that is often used for the purposes of classification, where the output of said classification is dichotomous [10]. It does so by finding the relationship between two data factors and then subsequently using this relationship to predict the value of one, based on the other, by utilising the logit function. A key benefit of using logistic regression is that it is not bound by the same assumptions as other models in the generalised linear model (GLM) family, specifically: a dependence on there being a linear relationship between one's dependent and independent variables; a dependence on the residuals in the data being normally distributed; and an assumption that the data in question is homoscedastic in nature [24]. For the purposes of this study, the statsmodel library was used to implement logistic regression, and the p-values, F1-score, and AUC-ROC were used to measure the performance of the model created.

### 2.3.3 Support Vector Machine

Support vector machine (SVM) is a supervised learning method that utilises maximum margin classifiers in order to separate data sets into classes by maximising the space between them [11]. They perform very well on small datasets, datasets with a high number of dimensions, are suited for imbalanced datasets, and are able to handle multicollinearity by use of an in-built regularisation process [11]. However, they also rely on support vectors, which are values at the border of a margin. These margins can be largely distorted by outliers, which can reduce the overall effectiveness of their classifications [11].

In this study, we used C-Support Vector Classification (C-SVC), which belongs to the family of SVMs. Using Bayesian optimisation, we found that the optimal hyperparameters for our model were a "C"-value of 1.0, a "gamma" of 0.1, and a kernel that utilised the radial basis function (RBF). "C" is a regularisation parameter that adjusts the trade-off between minimising the error and maximising the margin, gamma controls the influence of a single training example and the kernel describes the function used for transforming the data before linearly separating it. SVMs are computationally expensive and thus do not scale well for large datasets [11]. Therefore, we implemented C-SVM with a random subset of the data with 10,000 sample points.

### 2.3.4 Random Forest

Random Forest is a supervised learning method, capable of handling both classification and regression problems, that is based on a single fundamental concept — the wisdom of crowds. It involves the use of a large number of individual decision trees that operate as an ensemble [8]. It performs classifications by summing up each of the individual classifications output by each of the constituent trees within the "forest", and the class with the highest number of votes is the model's overall prediction [8]. For this study, a Random Forest classifier, whose optimal hyperparametric values were ascertained with GridSearchCV [18], was trained on a sample size of 125000 with 50 separate trees, a maximum tree depth of 9, and the quality of the splits was measured using Gini Impurity [16]. It should also be noted that, in order to avoid too much computation, bootstrapping or "bagging" was utilised so that only a random subset of our data was used to train each individual decision tree. In addition, it should be noted that, unlike with the distance-based algorithms tested in this study, scaling was not utilised, as a tree-based model would not have benefitted from it.

### 2.3.5 XGboost

XGBoost is a decision tree-based ensemble machine learning algorithm under the gradient boosting framework [20]. It was designed to be highly efficient, and records show that it shows better results than other models when it comes to supervised learning on structured data [22]. Similar to Random Forest, it works by training multiple decision trees on subsets of the data and then combining their predictions into a final one. However, the difference between the two is that while Random Forest allows each tree to work in parallel and then combines prediction, XGBoost works in a sequential manner, which means that the outcome of each tree depends on the previous one. While Random Forest is a Bagging method, XGBoost is a Boosting method, utilising the gradient descent architecture [23]. In our case, the model was trained and then tuned using GridSearchCV with a sample size of 100,000. The observed optimal parametric values were a learning rate of 0.35, the minimum loss reduction required to make a further partition on a leaf node of the tree was set at 0.1, and the evaluation metric was based on the binary classification error rate.

## 3. EXPERIMENTS & RESULTS

### 3.1 Dataset

The dataset used is a public dataset from the United States government. It contains information about crimes in the city of Chicago between the years 2001 and 2023, inclusive [1]. The dataset is a 1.17 Gigabyte CSV file with 7724493 records and 22 columns. The columns are of both a categorical and numerical nature and contain crime details such as the type of crime, dates, localities, and additional identifiers.

### 3.2 Results

#### 3.2.1 Overview

All five of our models were trained on the same training set with varying sample sizes, differing combinations of features selected, preprocessing techniques utilised, and algorithm-specific values chosen for hyperparameters on each model. It should be noted that the values for said hyperparameters were chosen with the aid of either Bayesian Optimisation Search or Grid Search, depending on the model. The training set as a whole consisted of 7,724,493 records, which was split 80:20 (80% for training data, 20% for test data), and any imbalances in the data with respect to the "Primary Type" feature were resolved using a combination of oversampling (SMOTE) and random undersampling.

#### 3.2.2 Classification Results

The results of these models, when attempting to predict theft in Chicago, can be seen in Table 1. For each model, both the F1-score and AUC-ROC were calculated. The model that appeared to perform the best, with regards to these metrics, was the XGBoost model, which trained on a sample size of 100,000 and achieved an F1-score of 0.86 and an AUC-ROC of 0.87, whereas the model that appeared to perform the worst was the logistic regression model, that trained on a sample size of 7,724,493 and achieved an F1-score of 0.52 and an AUC-ROC of 0.65.

**Table 1. Results of Five Different Models' Attempts at Predicting Theft in Chicago.**

| Algorithm Used | Features Chosen | Techniques Utilised | Sample Size | F1-Score | AUC-ROC |
|---|---|---|---|---|---|
| K-Nearest Neighbours | Beat, Block, Community Area, Domestic, Hour, Latitude, Longitude, Location Description, Month, Ward, Weekday, X Coordinate, Year | Grid Search, SMOTE | 125,000 | 0.74 | 0.74 |
| Logistic Regression | Arrest, Domestic, Block, Community Area, Hour, Month, Ward, Week Number | SMOTE | 7,724,493 | 0.52 | 0.65 |
| Random Forest | Beat, Block, Community Area, Domestic, Hour, Latitude, Longitude, Location Description, Month, Ward, Weekday, X | Grid Search, SMOTE | 125,000 | 0.75 | 0.75 |

| | Coordinate, Year | | | | |
|---|---|---|---|---|---|
| Support Vector Machine | Beat, Block, Community Area, Domestic, Hour, Latitude, Longitude, Location Description, Month, Ward, Weekday, X Coordinate, Year | Bayesian Optimisation Search, SMOTE | 10,000 | 0.72 | 0.7 |
| XGBoost | Domestic, Location Description, Hour, Is_AV, Is_DR, Is_ST, Is_TR, Month, Weekday (_Monday - _Sunday), X Coordinate, Y Coordinate, Year | Grid Search, SMOTE | 100,000 | 0.86 | 0.87 |

### 3.2.3 Feature Contributions

The levels of contribution for each feature in our models were measured using Shapley Additive Explanations (SHAP) [19], and the feature with the highest contribution for each model was noted (see Table 2). As is evident from Table 2, the majority of the features that provided the highest contributions to predictions were of a spatiotemporal nature. Interestingly, the features that appeared to provide some of the higher contributions were of a geographical nature: "Block" and "Location Description", although it cannot be ignored that the day of the week "Weekday_Friday", a time-based feature, also played an important role in contributing to the overall results of certain predictions.

Table 2. Highest Contributions for Each Model Based on Shapley Additive Explanations.

| Algorithm Used | Most Contributing Feature | Average Shapley Value |
|---|---|---|
| K-Nearest Neighbours | Domestic | 0.08 |
| Logistic Regression | Block | 2.0 |
| Random Forest | Location Description | 0.14 |
| Support Vector Machine | Domestic | 0.17 |
| XGBoost | Weekday_Friday | 0.55 |

## 4. DISCUSSION

### 4.1 Findings

Our research has shown that it is indeed possible to predict specific types of crime in the city of Chicago with use of artificial intelligence and a suitable dataset. The results outlined above show that a focus on spatiotemporal features (time and location) is a valid approach for predicting crimes of opportunity, such as theft, further corroborating the theories of both crime pattern theory [7] and "defensible space" [6], that place an importance on such features.

In the context of general crime prediction, the studies upon which this study was built achieved significant results when attempting to predict theft in China [4], South Korea [2], and the US [5]. We believe that the findings from this study indicate a certain degree of universality in how and why people from different countries, or even geographical regions, commit certain crimes, as the results achieved were consistent to a degree with the results achieved in the aforementioned studies.

And lastly, with regards to transparency and interpretability, we have shown that the use of Shapley Additive Explanations (SHAP) provides an informative insight into the factors that influence certain predictive models, when attempting to predict theft, further corroborating the theory put forward by [4], in which they suggested the use of an XGBoost model, paired with a SHAP tree to accomplish this same task.

### 4.2 Limitations & Weaknesses

With regard to limitations in our research, the dataset chosen is worth noting. Though the spatiotemporal features provided within were sufficient to construct a model to adequately predict theft, a more detailed and balanced dataset could have provided us with the information required to make more accurate predictions, such as information about the principal offender (age, ethnicity, profession, criminal history, etc.). Were an additional study to be carried out to attempt to improve predictions in the city of Chicago, we believe that it could benefit from a more informative dataset. A lack of imbalance in future data could also reduce the preprocessing time that was required to resolve the imbalance in question, although crime is an inherently imbalanced thing, and as such, it may be difficult to acquire a dataset that fits the aforementioned criteria.

It is also worth noting that, though significant results have been achieved with regard to prediction in the city of Chicago, this does not indicate whether or not the approaches used in this study are applicable to other countries or even other major cities within the US. As such, further research should be carried out to confirm whether or not this approach achieves the same results across varying locations.

### 4.3 Future Considerations

Some approaches that could be taken to improve the models created in this study are: as mentioned above, to find a dataset that contains more information on the individuals involved in the crimes committed; to utilise a deep learning model in order to ascertain whether or not this has any significant impact on our prediction results; to apply additional imputation and synthesis techniques to our existing dataset in order to determine whether or not any improvements could be made to it; and lastly, to compare the results achieved when predicting other, less commonly occurring types of crime, such as ritualism.

### 4.4 Conclusion

In conclusion, we managed to predict theft using only spatial and temporal features while providing a better insight into the calculations happening behind the models, making the problem more transparent and interpretable. This paper also confirms the effectiveness of XGBoost in the context of crime prediction as it was stated in previous research. Finally, although the model seems to show promising results, more information regarding the crimes could potentially improve our model. Plus, we do not know if it would still be the same with data from another city.

### SOURCE CODE

All source code referred to herein is publicly available at the following link: https://github.com/deborahdjon/Chicago_Crime_Prediction.